\documentclass[runningheads]{llncs}
\usepackage{geometry}
\usepackage{microtype}
\usepackage{graphicx}
\usepackage{subfigure}
\usepackage{booktabs}

\usepackage{amsfonts}
\usepackage{amsmath}
\usepackage{graphicx}
\usepackage[colorlinks=true, allcolors=blue]{hyperref}
\usepackage{tikz-cd}
\usepackage{caption}

\usepackage{amssymb}
\usepackage{dsfont}
\usepackage{nicefrac}

\usepackage{amsmath}
\usepackage{amssymb}
\usepackage{mathtools}

\usepackage[capitalize,noabbrev]{cleveref}

\newcommand{\cE}{\mathcal{E}}

\newcommand{\R}{\mathbb{R}}

\newcommand{\E}{\mathbb{E}}
\newcommand{\Ah}{\widehat{A}}
\newcommand{\cM}{\mathcal{M}}
\newcommand{\cU}{\mathcal{U}}
\newcommand{\cC}{\mathcal{C}}
\newcommand{\cA}{\mathcal{A}}
\newcommand{\cX}{\mathcal{X}}

\newcommand{\cN}{\mathcal{N}}

\newcommand{\cL}{\mathcal{L}}

\newcommand{\bx}{\mathbf{x}}
\newcommand{\by}{\mathbf{y}}
\newcommand{\Var}{\text{Var}}
\newcommand{\noise}{\text{noise}}

\usepackage[textsize=tiny]{todonotes}

\begin{document}

\title{Explaining, Evaluating and Enhancing Neural Networks' Learned Representations}

\author{Marco Bertolini \and
Djork-Arn\'e Clevert \and
Floriane Montanari
}
\authorrunning{Bertolini et al.}

\institute{Bayer AG, Machine Learning Research, Berlin, Germany\\
\email{\{marco.bertolini, djork-arne.clevert, floriane.montanari\}@bayer.com}
}

\maketitle
\begin{abstract}

Most efforts in interpretability in deep learning have focused on (1) extracting explanations of a specific downstream task in relation to the input features and (2) imposing constraints on the model, often at the expense of predictive performance.
New advances in (unsupervised) representation learning and transfer learning, however, raise 
the need for an explanatory framework for networks that are trained without a specific downstream task. 
We address these challenges by showing how explainability can be an aid, rather than an obstacle, towards better and more efficient representations.
Specifically, we propose a natural aggregation method generalizing attribution maps between any two (convolutional) layers of a neural network.
Additionally, we employ such attributions to define two novel scores for evaluating the informativeness and the disentanglement of latent embeddings. Extensive experiments show that the proposed scores do correlate with the desired properties. We also confirm and extend previously known results concerning the independence of some common saliency strategies from the model parameters. Finally, we show that adopting our proposed scores as constraints during the training of a representation learning task improves the downstream performance of the model.
\end{abstract}

\section{Introduction}
\label{s:intro}

The steadily growing field of representation learning \cite{GRL_Hamilton,NLPReviewRL} has been investigating possible answers to the question “how can we learn good representations?”. This has led to the rise of self-supervised learning and transfer learning \cite{TL_survey}. In practice, the quality of a representation is measured by the performance on the downstream task of interest. This can lead to a proliferation of models that potentially overfit the few benchmark datasets typically used as downstream tasks.

In this work, we would like to step away from such traditional evaluations and instead come back to the seminal work by \cite{Bengio2013RepresentationLA}, questioning “what makes one representation better than another?”. Our proposal arises from the field of explainable AI (XAI) \cite{gunning2017explainable,Samek2019,dovsilovic2018explainable,BARREDOARRIETA202082}. In XAI, the primary goal is usually to provide an interpretation of a model’s predictions either in terms of the input features or in terms of some user-defined high-level concepts \cite{Kim2018InterpretabilityBF}. Here as well, the field has suffered from the same endpoint-bias propelling model development, as XAI methods are in most cases designed to extract explanations for a specific downstream task.  This is clearly a limitation: such methods cannot be directly applied to ``explain'' embeddings learned in a self-supervised manner. 

In this paper, we develop an XAI framework that does not rely on any downstream network or task, and we then show how we can adopt it to evaluate the quality of an embedding. This requires a generalization of the concept of explanation: what we try to achieve here is to visualize and compare embeddings learned by any neural network.

Our explainability framework is local (we provide a method to infer explanations for a single data input), independent of any downstream task, and flexible (the framework is model-agnostic even though most of our examples and experiments are given for CNNs applied to image analysis). The scores we derive are global and can interpret a whole embedding vector.

\subsection{Related Work}
\label{ss:related}

To visualize global concepts learned by trained CNNs, one can employ techniques such as filter visualization \cite{Zeiler2014VisualizingAU,Mahendran2015UnderstandingDI,Simonyan2014DeepIC} or activation maximization techniques. Those aim at extracting visual patterns that maximize an internal neuron or predefined direction in the weight space \cite{Mordvintsev2015InceptionismGD,Nguyen2019UnderstandingNN,46832}. In this context, \cite{Szabo2020VisualizingTL} visualizes hidden neurons to gain insights on the property of transfer learning.

In the subfield of local explanations, saliency methods are commonly used to generate and visualize attributions for specific inputs. In our work, we compare three common strategies: Vanilla Gradients \cite{Baehrens2010HowTE,Simonyan2014DeepIC}, Activations$\times$Gradients (which generalizes Gradient$\times$Input \cite{Shrikumar2017LearningIF} to any hidden layer), and Grad-CAM \cite{Grad-CAM}. Although our techniques can be similarly applied to any saliency method, the three above suffice to show our proof of concept, and avoid the computational overhead that comes with, for example, Integrated Gradients \cite{Sundararajan2017AxiomaticAF}. \cite{Chen_2020_WACV} proposed a variation of the Grad-CAM method for Embedding Networks. Their approach however fundamentally differs from ours, as the authors explicitly utilize a downstream loss function for computing attributions. The authors of \cite{8999235}, instead, extend saliency methods to
visualize relevant features in latent space for generative models.

An alternative strategy to increase model interpretability is to restrict the architecture to intrinsically interpretable ones \cite{DoshiVelez2017TowardsAR} or to impose other constraints during the training process \cite{Zhang2018InterpretableCN,Zhou2019InterpretingDV,Henderson2021ImprovingMG,DBLP:journals/corr/abs-2007-01760}. One frequently used constraint is to enforce the disentanglement of the learned representation. However, this increased interpretability through constraints often comes at the cost of predictive performance \cite{Zhang2018InterpretableCN,DBLP:journals/corr/abs-2007-01760}. We will show that this need not be the case: our XAI-derived scores do help the model achieve higher predictive power.

Several works \cite{Alain2017UnderstandingIL,Raghu2017SVCCASV,Bau2017NetworkDQ,Szegedy2016RethinkingTI,Engel2018LatentCL} demonstrated that neural networks learn high-level concepts in the hidden layers.  In particular, \cite{Kim2018InterpretabilityBF} showed that (1) such concepts can be represented as directions in the latent space of hidden layers, and (2) deeper layers tend to learn more complex concepts as opposed to shallow layers. The disadvantage of their approach is that such meaningful directions are learned with the aid of additional labelled data. Our proposal generalizes this basic idea to learned embeddings by aggregating the concepts, learned in hidden layers, that contribute to activate the target representation. Our approach is partly motivated by the work \cite{Fong_2018_CVPR}. The authors show that, often, single filters fail to capture high-level concepts, and that filter aggregation is instead required.

\subsection{Contributions}
We can summarize our main contributions as follows:
\begin{itemize}
    \item We propose a simple and flexible aggregation scheme for explaining and visualizing embeddings of (convolutional) neural networks (Section \ref{s:aggreg}). To the best of our knowledge, the present work is the first systematic attempt to do so without any consideration of a downstream task or loss. 
    \item We propose in Section \ref{s:scores} metrics to quantify the quality of these embeddings.
    \item We show that these metrics correlate with desired embedding properties (specifically, informativeness and disentanglement) (Section \ref{ss:ns_attr_comp}, 
    \ref{ss:rand_layers}, \ref{ss:conciseness}) and detect effects like dataset drift (Section \ref{ss:drift}).
    \item We show that training with the proposed metrics as constraints improves the quality of the embeddings (Section \ref{ss:enhance}).
\end{itemize}

\section{Visualizing Explanations of Embeddings through Aggregation}
\label{s:aggreg}

\begin{figure*}
    \centering
    \includegraphics[width=1.0\textwidth]{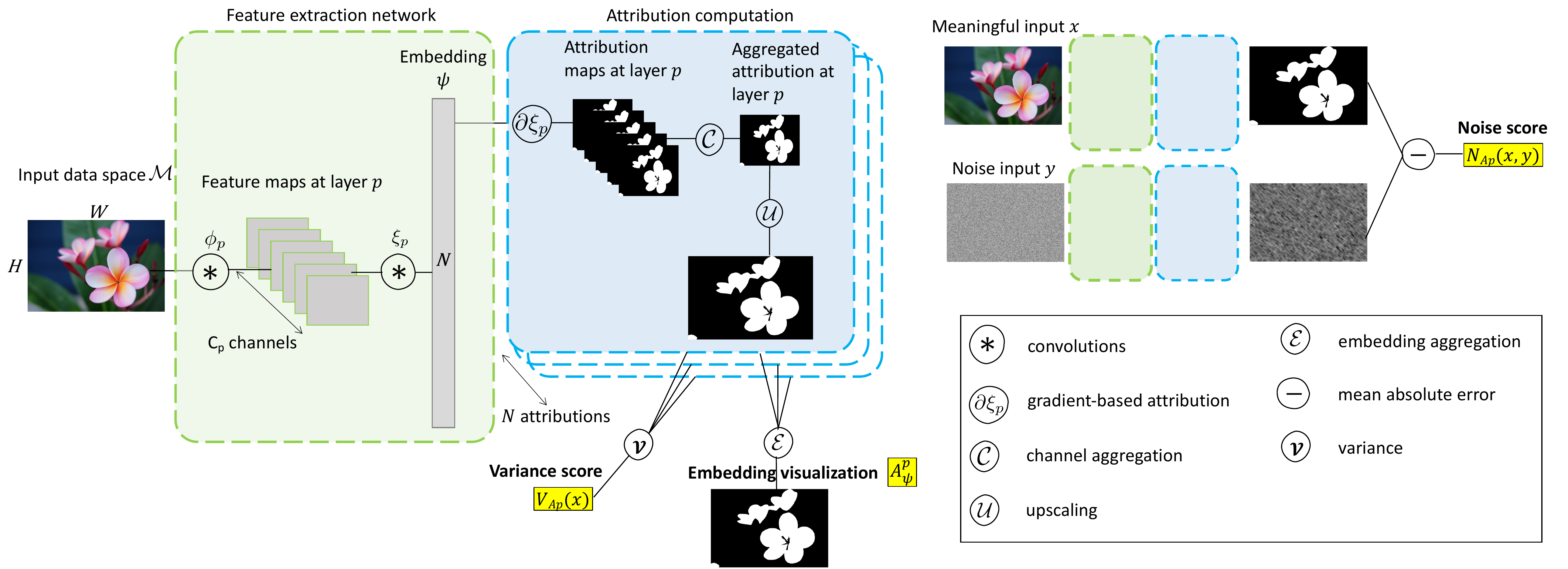}
    \caption{Schematic description of the embedding explanation framework, including how the noise and variance scores are calculated.}
\label{f:genidea}
\end{figure*}

In this section, we present a simple and intuitive procedure for 
visualizing latent embeddings of a neural network. 

Let $\Psi$ denote a representation learning task, that is, $\Psi: \cM \rightarrow \R^{N}$, where $\cM\simeq \R^d$ is the manifold 
describing the input data. For $\mathbf{x}\in\cM$, $\Psi(\mathbf{x})$ denotes the target embedding, which is supposed to be a meaningful, lower-dimensional representation of the data. 
Relevant examples are, for instance, the bottleneck layer of an autoencoder network, or
the output of the last convolutional layer in a image analysis NN.
To fully characterize the concepts captured in the learned representation,
and to mitigate the influence of the input features on the attribution map (as we shall see in Section \ref{ss:ns_attr_comp}), it is insightful
to consider attributions that probe latent dimensions. 
Thus, we describe our task as $\Psi: \cM \xrightarrow{\phi_p} \cM_p \xrightarrow{\xi_p} \R^N$,
where $\cM_p$ is the input space of the $p^{\text{th}}$ layer in the network (in this notation, $\cM_0 \equiv \cM$).
According to our assumption of a convolution structure, $\cM_p \simeq \R^{C_p\times d_p}$, 
where $C_p$ is the number of channels and $d_p$ is the spatial dimension of the latent representation.

\subsection{Attributions for Embeddings}

To generate explanations for embeddings without any reference to 
a downstream task, we regard each target embedding component $\Psi(\bx)_i$, for $i=1,\dots,N$,
as a stand-alone output node. Consequently, we interpret its corresponding saliency 
attributions $\partial \Psi(\bx)_i / \partial \bx$ as feature importance
for the activation of the corresponding latent component. This approach can be
straightforwardly extended to any intermediate layer, where we regard the attribution values 
$\partial \xi_p(\bx)_i / \partial \mathbf{\phi}_p(\bx)\in \cM_p$ 
as importance weights for the activation of the $i^{\text{th}}$ representation
component with respect to the $p^{\text{th}}$ layer neuron activations $\partial \mathbf{\phi}_p(\bx)$. 

In general, $\dim \cM_p \neq \dim \cM$, as the above procedure will
generate, for each $i$, a map for each of the $C_p$ channels in the feature layer $p$. 
To generate a unique map for the attribution, we
aggregate the maps $\partial \xi_p(\bx)_i / \partial \mathbf{\phi}_p(\bx)$ along the channel dimension of layer $p$
by means of a map $\cC: \cM_p \mapsto \R^{d_p}$, such that
$\cC:\partial \xi_p(\bx)_{i=1,\dots,N} / \partial \mathbf{\phi}_p(\bx) \mapsto 
(\Ah_\Psi^p)_{i=1,\dots,N}\in \R^{d_p\times N}$.

In most representation learning applications $N \sim(10^2, 10^3)$, 
and it is unfeasible to examine saliency maps for \textit{all} latent dimensions $i$.  
A simple solution to generate one single map per input sample is to aggregate 
the component-wise attribution maps via $\cE: (\Ah_\Psi^p)_i \in \cM \mapsto \Ah_\Psi^p \in \cM_p$. 

It is desirable to generate the final attribution in the original input space, 
in order to formulate hypothesis about explanations on the input features. 
However, it can happen that $d_0 > d_p$, for instance due to the presence of pooling layers. When this is the case, we apply an
upscaling map $\cU : \R^{d_p} \rightarrow \R^{d_0}$. 
Given the property of a convolutional layer that the value of each superpixel depends on its receptive field, 
the simplest choice for $\cU$ is to consider the map in $\R^{d_p = H_p \times W_p}$ as a low-resolution map in $\R^{d=H \times W}$.
This is the choice we implemented in all our examples and experiments.\footnote{
Another upsampling scheme, for instance, involves performing a transposed convolution of the lower-dimensional matrix
with a fixed Gaussian kernel to obtain a full-resolution map \cite{Liznerski2020ExplainableDO}.}

\begin{figure*}
    \centering
    \includegraphics[width=0.95\textwidth]{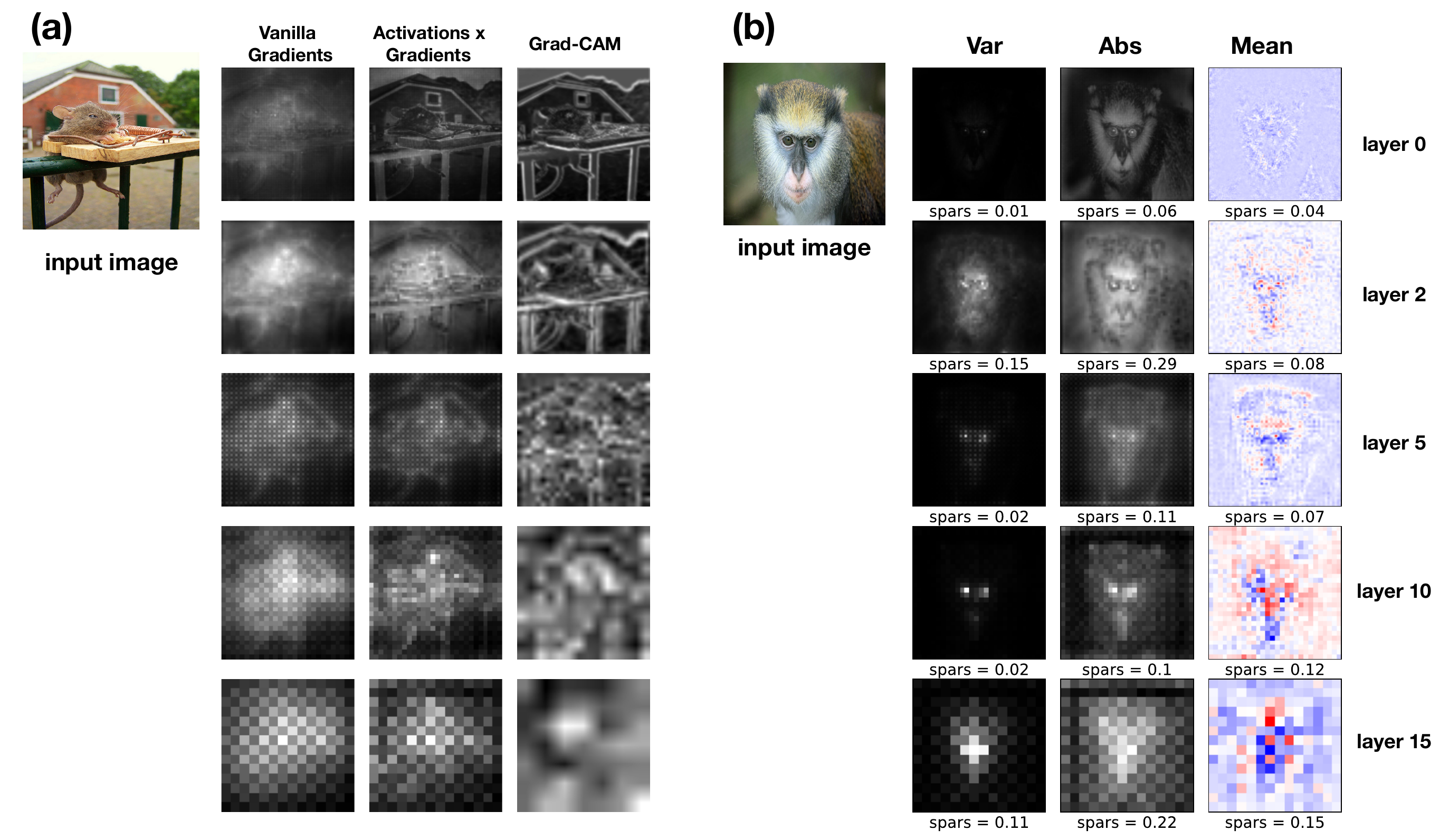}
    \caption{Example of embeddings for (a) different attribution schemes 
    ($\cC=\cA_{\text{abs}}$, $\cE=\cA_{\text{var}}$) and 
    (b) aggregation strategies (Activations $\times$ Gradients). Below each map we report its corresponding sparsity score,
    given as the average value over the normalized 2D attribution map.
    We report in the appendix further examples and combinations of
    attribution and aggregation schemes.}
\label{f:attribs}
\end{figure*}

Putting all together, given a learning task $\Psi$, we define the attribution method $A_\Psi^p$
as a map
\begin{align}
\nonumber
\small
    \cM \xrightarrow[\text{forw.}]{\Psi} \R^N \xrightarrow[\text{back}]{\partial \xi_p} \cM_p^N
    \xrightarrow[\text{aggr.}]{\cC} \R^{d_p \times N}
    \xrightarrow[\text{aggr.}]{\cE} \R^{d_p} \xrightarrow[\text{upscale}]{\cU} \cM~.
\end{align}
We depicted this chain of operations in Figure \ref{f:genidea}.
In the preceding discussion, to keep notations simple, we 
explicitly considered vanilla gradients $\partial \xi_p$ as the reference saliency method. This 
however can be replaced with any available saliency method. Figure \ref{f:attribs} shows, for two input samples, examples of embedding attributions for different
saliency and aggregation strategies. We notice that for deeper layers the attributions focus on some high-level concepts (e.g., eyes and nose for Figure \ref{f:attribs}b).

Following \cite{Bau2017NetworkDQ,Alain2017UnderstandingIL,Kim2018InterpretabilityBF}, 
latent dimensions contain meaningful directions that can be learned through linear classifiers and that correspond to humanly-understandable concepts. 
The attribution maps $(\Ah^p_\Psi{})_{i,c}: \cM \rightarrow \R^{d_p}$, 
where $c=1,\dots,C_p$, is to be understood
as quantifying the concept(s) learned by the model in 
the $c^{\text{th}}$ filter of the $p^{\text{th}}$ layer. 
Thus, in our approach, we regard the channel aggregation map $\cC$ as a ``concept'' aggregation map. 
On the other hand, aggregation along the latent embedding dimension
is dictated by the need to ``summarize'' explanations across the whole
representation. 
Given a downstream task $\Theta : \R^N \rightarrow \R$, however, each embedding dimension $\Psi_i$ acts simply as an input feature. 
With regard to explainability, we interpret the explanations for $\Psi_{i=1,\dots,N}$ as a basis of explanations for $\Theta$. 
Namely, explanations $A_\Theta$ for the task $\Theta$ would act as weights for the more elementary explanations $A_{\Psi_i}$.

\subsection{Aggregation Schemes}

Here we review possible aggregation strategies for both the channel and embedding dimensions. 
To formalize our discussion, we represent the set of attribution 
maps as a multiset $\cX$, that is, a set allowing for repeating elements. 
Both the channel ($\cC$) and embedding ($\cE$) maps can be thought of as an aggregation map
$\cA(\cX)$ over a multiset $\cX$.
We also propose a new 
aggregation strategy, given by $\cA = \Var[\cX]$, that yields sparser attribution values. 

\paragraph{\textbf{Mean aggregation.}}

The simplest 
aggregation method, 
is the \textit{mean} aggregator, $\cA_{\text{mean}}(\cX) = \frac1{|\cX|}\sum_{x\in\cX} x$. 
The mean aggregator captures the distribution of 
elements in a multiset \cite{Xu2019HowPA}, but not the multiset itself. 
For example, for $\cX_1 = \{-1,1,0,0\}$ and $\cX_2=\{0,0,0,0\}$
we obtain $\cA_{\text{mean}}(\cX_1) = \cA_{\text{mean}}(\cX_2) = 0$. 
Such aggregations ``with signs'' potentially suffer from the risk of signal cancellation. 

\paragraph{\textbf{Abs aggregation.}}
Another simple aggregator is given by the \textit{mean absolute} aggregator,
$\cA_{\text{abs}}(\cX) = \frac1{|\cX|}\sum_{x\in\cX} |x|$.
The mean absolute aggregator captures 
the distribution of elements in a multiset without signs. 
Consequently, the resulting attributions tend to be quite dense. In fact,
$\cA_{\text{abs}} =0$ if and only if all elements in $\cX$ are zero. 
This feature could be detrimental in our setting,
as sparsity is often a desired property in explainability, 
and especially challenging to achieve in 
post-hoc explanations.

\paragraph{\textbf{Variance aggregation.}}
To remedy this lack of sparsity, we introduce a novel aggregation scheme,
the \textit{variance} aggregator,
$\cA_{\text{var}} = \Var[\cX]$. 
The variance aggregator is somewhat complementary to the mean aggregator. In fact,
for $\cX_1 = \{-1,1,0,0\}$, we have $\cA_{\text{var}}(\cX_1) = \frac{2}{3}$
while $\cA_{\text{mean}}(\cX_1)=0$. On the other hand, for 
 $\cX_2=\{1,1,1,0\}$, $\cA_{\textit{var}}(\cX_2) = \frac{1}{4}$
while $\cA_{\text{mean}}(\cX_2)=\frac{3}4$.
Thus, the variance aggregator 
assigns higher values to multisets with a wider range of values
and penalizes biased multisets where all values concentrate around its
mean, regardless of its magnitude.
We derive in the Supplementary Information (SI) a simple
argument that shows, under a few basic assumptions, that the variance
aggregator yields a sparser map than the mean absolute aggregator. 
We provide evidence that our claim extends to real data by displaying
different aggregation methods in Figure \ref{f:attribs}b.
For each attribution map, we computed a measure of sparsity 
as the average value over the (normalized to the range $[0,1]$) 2D attribution map. 
For the mean aggregator, the sparsity is computed on the absolute value of the map.
As claimed, for all layers, the variance produces a sparser map
than the mean absolute aggregator.

\paragraph{\textbf{Comments and practical advices.}}

Finally, we provide practical guidance as to the combinations of aggregation schemes for the maps $\cC, \cE$. 
We find the combination $\cC = \cA_\text{abs}$, $\cE = \cA_{\text{var}}$ (and viceversa) especially insightful
to identify the prevailing features dominating the representation, while 
$\cE \equiv \cC = \cA_\text{abs}$ is helpful to detect features that are \textbf{not}
captured by the embedding (see Figure \ref{f:attribs}).
We find instead $\cE \equiv \cC = \cA_{\text{mean}}$ potentially useful when the embedding explanations are combined in a generalized chain rule with a downstream task $\Theta$ as
mentioned above, 
especially when it is desirable to distinguish feature contributing positively or negatively.
When $\cC \equiv \cE = \cA_{\text{var}}$, we observed that the map is often too sparse.

\section{Explainability-derived Scores}
\label{s:scores}

In this section, by employing the embedding explanations described in the previous section,
we derive useful metrics to quantify desired properties of a learned latent representation: (1) informativeness,
the ability of a latent representation to capture meaningful features of the input sample, and
(2) disentanglement, the property quantifying the degree to which 
latent features are independent of one another. 
We will show that our proposed scores correlate with the desired properties as well as with the more standard approach of predictive performance on a downstream task.

\subsection{Informativeness: the Noise Score}
\label{ss:noise_score}
The motivating idea of the noise score is that attributions for an informative
representation should, on average, substantially differ from attributions generated from a less meaningful representation. 
Explicitly, let $x\in \R^{d}$ be a (meaningful) input image, and let $x_{\noise}\in \R^{d}$ be such that $x_{\text{noise}} \sim U(0,1)^{d}$. 
We define the \textit{noise score} 
for representation task $\Psi$ and layer $p$ as
the Mean Absolute Error (MAE) between (normalized) attributions of a meaningful data input and a noise-generated input 
\begin{align}
\label{eq:noisescore}
    N_{A^p_\Psi}(x, x_\noise) = \overline{|A^p_\Psi(x) - A^p_\Psi(x_\noise)|}~.
\end{align}
We schematically depicted the above definition in Figure \ref{f:genidea}.
The attribution maps in \eqref{eq:noisescore} are normalized
to the range $[0,1]$ to ensure
that the score does not merely reflect the magnitude of
the respective attribution values. With such a normalization, we have $\sup N_{A^p}(x, y) = 1$,
which occurs if $A^p_\Psi(x) -  A^p_\Psi(y) = \pm1$,
while $\inf N_{A^p}(x, y) = 0$, which occurs when $A^p_\Psi(x) \equiv A^p_\Psi(y)$. 

We will show in the next section that \eqref{eq:noisescore} distinguishes
different representations for which we know their relative informativeness.
However, it is often convenient to have a global reference value $N^0$,
to establish an intrinsic metric for \textit{informativeness}.
Since we do not know
the underlying distribution of $A^p_\Psi$, we can sample 
a benchmark value
as follows:
\begin{align}
\label{eq:benchscore}
    N^0_{A^p_\Psi} = \overline{|A^p_\Psi(y_\noise) - A^p_\Psi(x_\noise)|}~,
\end{align}
where $y_{\text{noise}} \sim U(0,1)^d$.

\paragraph{\textbf{The score curve.}}
Given a representation task $\Psi$, 
we can compute 
\eqref{eq:noisescore} and \eqref{eq:benchscore} for each intermediate layer $p$ of the network. The resulting \textit{noise score curve} provides a more complete
picture of how effectively the different layers organize information in the target
representation.

\subsection{Conciseness: the Variance Score}
\label{ss:var_score}
A second beneficial property of a representation is disentanglement \cite{Eastwood2018AFF,Ridgeway2018LearningDD,Do2020TheoryAE}. 
Disentanglement, in the sense of 
decomposability \cite{2016arXiv160603490L}, implies that the latent features are independent of one another. 
As a consequence, we also intuitively expect the attributions for each of the latent features not to overlap among each other significantly.
Of course, a pair of latent features $\Psi_{1,2}$ can be independent but have very similar attribution maps for a given input, as they might encode independent concepts that happen to overlap in the given sample. 
Still, we expect the majority of the input space to be represented by
some high-level concepts, and a decomposable representation should cover 
a large portion of it. 
Since our choice of measuring disentanglement is correlated
but differs from the more standard definition, we coin a new term:
the \textit{conciseness} of a representation is a measure of the variance of attribution maps across the latent features.

Explicitly, let $x\in \R^d$ be a (meaningful) sample input. 
We define the \textit{variance score} for the representation task $\Psi$
at layer $p$ as the mean variance of the channel-aggregated (but not embedding-aggregated)
attribution maps $(\Ah^p_{\Psi})_i(x)$, i.e., 
\begin{align}
\label{eq:varscore}
    V_{A^p_\Psi} (x) = \overline{\Var \Ah^p_\Psi (x)}~.
\end{align}
Figure \ref{f:genidea} illustrates how \eqref{eq:varscore} relates to the
aggregation strategies of Section \ref{s:aggreg}.
The above definition enables a direct comparison of embeddings: a representation $\Psi$ is more concise than representation $\Psi'$ if $V_{A_{\Psi}^p} (x) > V_{A_{\Psi'}^p} (x)$. 

In order to determine the intrinsic conciseness of a representation, we recall Popoviciu's inequality \cite{Popoviciu} for a bounded random variable $Z$, 
which reads $\text{Var}Z \leq \frac{(\sup Z - \inf Z)^2}{4}$,
yielding in our case $V_{A^p_\Psi}(x) \leq 1/4$, assuming that each dimension map $(\Ah^p_{\Psi})_i$
is normalized in the range $[0,1]$. 
The only random variable $Z$ whose variance assumes the maximum value is 
such that $Z = 0, 1$  each with probability $\frac12$. In terms of our map, this would represent that a ``superpixel'' is either considered on/off 
in each representation component (with equal distribution). Note that this is independent from the number of drawings, or in our case the embedding size. 
Hence, \eqref{eq:varscore} is suitable for comparing representations of any size,
as we will do in Section \ref{ss:conciseness}.

\section{Experiments and Results}
\label{s:exps}

In this section we describe our experiments and illustrate the insights
we gain from the metrics we defined in the previous section.
We apply our techniques to three widely used architectures: 
Inception V3 \cite{Szegedy2016RethinkingTI}, ResNet18
\cite{DBLP:journals/corr/HeZRS15}
and AlexNet \cite{Krizhevsky2014OneWT}.\footnote{Resnet18 has a 
BSD 3-clause license, Inception V3 has an MIT license, AlexNet has an Apache V2 license.} 
For each of these, the target representation 
will be the output of the last convolutional (or eventually pooling) layer.

We conduct several tests to assess the sensitivity of our scores to  informativeness, disentanglement and predictive performance.
Specifically, we compare the values of our scores for a given task and model to the one computed from networks with random weights.
In addition of being natural benchmarks for the desired properties listed above, 
randomly initialized networks do 
correspond to non-trivial representations \cite{Ulyanov2018DeepIP}.
Thus, our scores can be interpreted in this case as probing the model architecture as a prior to the 
learning task \cite{Saxe2011OnRW,Alain2017UnderstandingIL}. 

\begin{figure*}[t!]
  \centering
  \includegraphics[width=1.0\textwidth]{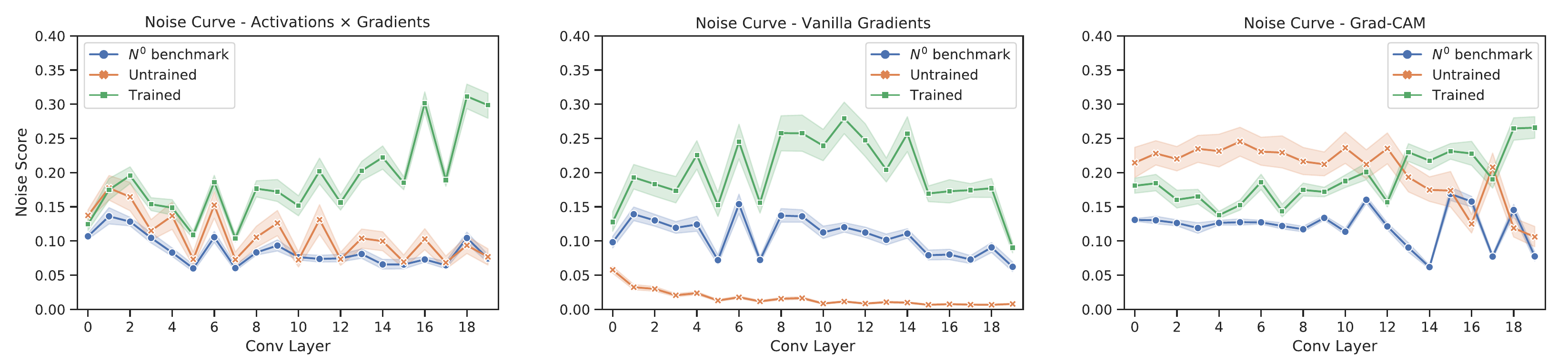}
  \caption{Average noise score curves (over 50 input samples) for ResNet18 trained on \texttt{ImageNet} \cite{Russakovsky2015ImageNetLS} for different attribution
  strategies and $\cC=\cE=\cA_{\text{abs}}$.}
  \label{f:scorecurve_aggr}
\end{figure*}

\subsection{Detecting Model's Parameter (in)dependence}
\label{ss:ns_attr_comp}

We begin by comparing the noise score \eqref{eq:noisescore} 
for a fixed network architecture and
fixed task $\Psi$.
Score curves for different attribution strategies are
presented in Figure \ref{f:scorecurve_aggr}.
For Activations $\times$ Gradients, the noise score fail to
distinguish between the trained and the untrained network in
early layers. This agrees with the findings of 
\cite{Adebayo2018SanityCF}, where the authors 
show that upon visual inspection the defining structure
of the input is still recognizable for a network with randomly initialized weights. 
Our analysis shows that this does not occur for explanations in deeper layers, as
the two curves separate further as we go deeper in the network. Thus, while 
Activations $\times$ Gradients is potentially misleading in shallow layers (due to the bias introduced by the input values), 
it is a meaningful attribution method in deeper layers (we defer explicit examples of such maps to the SI).

For Vanilla Gradients, we notice instead that the two curves (trained and untrained) are well separated even in the
early layers. This confirms the analysis of \cite{Adebayo2018SanityCF}, where the authors found
a significant discrepancy between the corresponding saliency maps.
The difference is however less conspicuous in comparison to the benchmark curve computed from \eqref{eq:benchscore}. This shows that Vanilla Gradients is more sensitive
to weight parameters than input parameters (we recall that the
$N^0$ benchmark curve \eqref{eq:benchscore} is computed for a trained network). This is not the case for 
Activations $\times$ Gradients, where the noise curves for the untrained network and the 
benchmark exhibit a very similar behaviour. 

Finally, we observe an overall poor performance of Grad-CAM, except 
in the very deep layers, where the noise score successfully distinguishes 
the trained network from both benchmarks \cite{Adebayo2018SanityCF}. This is consistent with the usual 
practice of considering the output of the last convolution layer for a Grad-CAM
visualization.

\subsection{Detecting Downstream Performance}
\label{ss:rand_layers}

\begin{figure*}[t!]
\begin{minipage}{\textwidth}
  \begin{minipage}[t]{0.49\textwidth}
    \centering
    \captionsetup{type=figure}
    \includegraphics[width=1.0\textwidth]{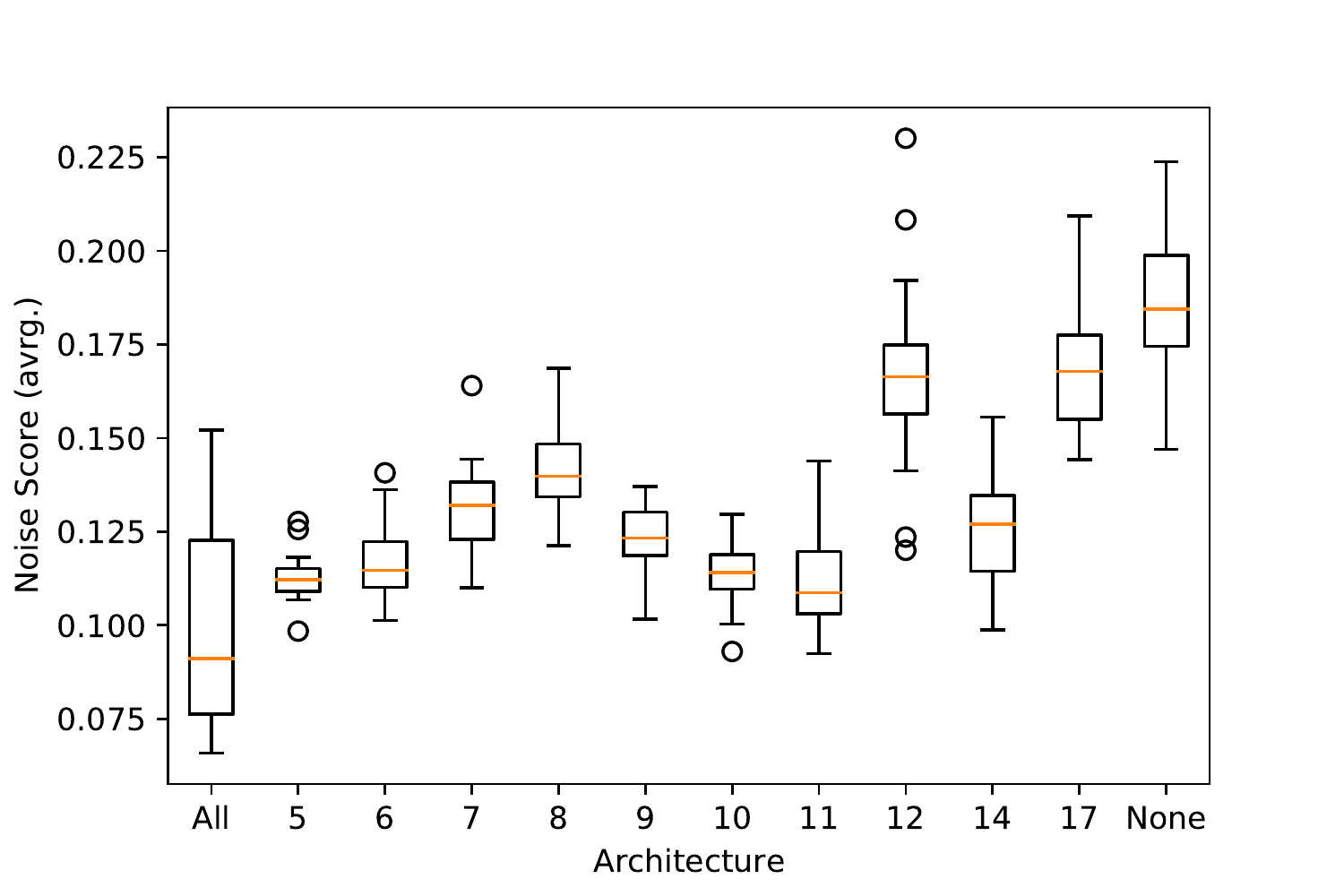}
    \captionof{figure}{Average noise scores ($25$ samples) for ResNet18 networks with layer-wise 
    randomly initiated weights.
    Each data point is obtained by averaging the noise score for each layer of the noise curve.
    $x$-axis values correspond to the re-initialized layer's number.}
    \label{f:rand_layers}
  \end{minipage}
  \hfill
  \begin{minipage}[t]{0.49\textwidth}
  \vspace{2mm}
    \centering
    \captionsetup{type=figure}
    \captionof{table}{Classification performance for single layer randomization.
    All=completely re-initialized network, None=full trained network.}
    \label{t:rand_perf}
    \small
 \begin{tabular}[b]{c|cc}
    \toprule
    Layer     & Top 1 Acc & Top 5 Acc \\
    \midrule
    All       & 0.0       &  0.7 \\
    5       & 0.0   &    1.1   \\
    6       &  0.2       & 0.9 \\
    7       & 0.7   &   1.8     \\
    8       &  0.3       &  2.8 \\
    9       &  1.8   &  4.2     \\
    10       &  0.2      & 0.3 \\
    11       & 0.2   &   0.3    \\
    12      & 63.7 & 85.1 \\
    14       & 4.3      & 10.1 \\
    17       & 30.0       &  54.3 \\
    None       & 74.2  &    91.8  \\
    \bottomrule
  \end{tabular}
  \vspace{2mm}
  \end{minipage}
\end{minipage}
\end{figure*}

The noise score \eqref{eq:noisescore} 
aims at quantifying the amount of information captured
by the target embedding. It is therefore expected to correlate with 
a given downstream task performance. 
We test this expectation by comparing classification performance
and noise scores of embeddings obtained through an independent layer randomization.
Explicitly, we fix the weights of all the layers to their
trained value except for one layer, whose weights we randomize. 
We report in Table \ref{t:rand_perf} the classification performance 
for the ResNet18 architecture, and in Figure \ref{f:rand_layers}
we depict the corresponding embeddings' noise scores, defined as the average value of the
corresponding noise curve.

Table \ref{t:rand_perf} shows that
layer
randomizations have different effects on the overall performance of the representation, as
some networks (12,17) are still predictive. 
In Figure \ref{f:rand_layers} we observe that embeddings which are still predictive
exhibits a higher average noise score.
The Spearman rank-order correlation coefficients between average noise score and downstream performance are $0.89$ (Top5) and $0.91$ (Top1), showing a very strong correlation between 
the noise score and the downstream performance of the
representations.

\subsection{Detecting Representations' Conciseness}
\label{ss:conciseness}

Increasing the dimension of a representation increments its
expressive power. However, large embeddings incur in the risk of noisy and redundant 
representations, which in turn have detrimental effects on the learning efficacy. 
\begin{figure*}[t!]
  \centering
  \includegraphics[width=1.0\textwidth]{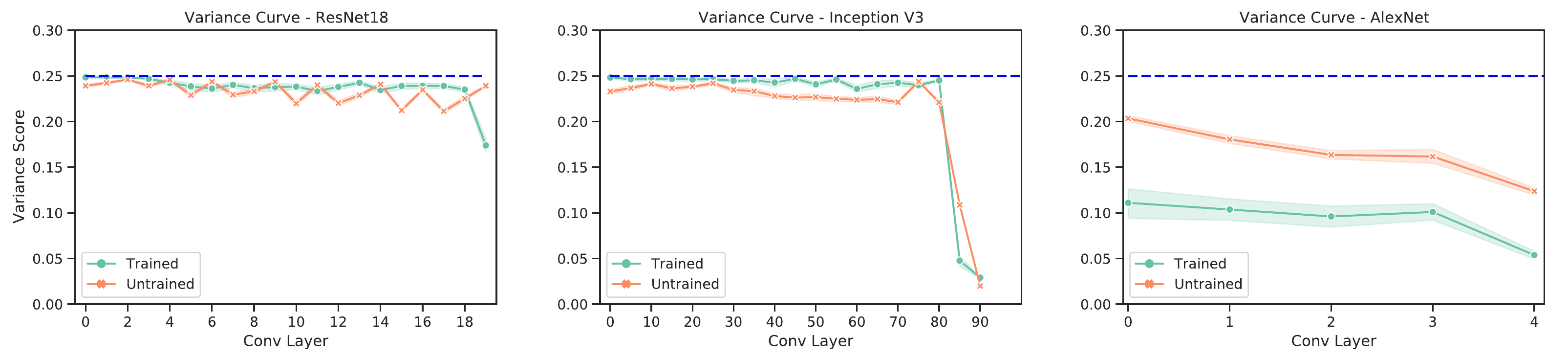}
  \caption{Variance curves for three popular networks trained on Imagenet. ($\cC=\cA_{\text{mean}}$, attr.= Activations $\times$ Gradients.) The blue dashed line depicts the theoretical maximum value.}
  \label{f:var_curves_NNs}
\end{figure*}

Here we provide evidence that
the variance score \eqref{eq:varscore} is suited to detect such pathology. 
Namely, we compute the variance curve for a given task (image classification
on the \texttt{ImageNet} dataset \cite{Russakovsky2015ImageNetLS})
and attribution scheme for different architectures. The corresponding variance curves are reported in Figure \ref{f:var_curves_NNs}. 
The attributions $\Ah_\Psi^p$ have been rescaled to the values $0,1$ (after normalization)
according to the threshold $0.5$.
The variance curves of the untrained models
indicate the conciseness prior of the pure model architecture, which is affected, for
example, by the choice of final non-linearity. 
In fact, we can assume that 
randomly initialized networks will tend to maximize representation conciseness, compatible
with the architecture prior. 
In our experiments, we observe
that the AlexNet architecture has a strong effect on the variance of the embedding
(in average around $75\%$ of the upper bound), while 
for Inception V3 and ResNet18 both curves approach the theoretically 
maximum value of $1/4$ (see Section \ref{ss:var_score}).

Next, we observe that for the trained Inception V3 and ResNet18, whose 
target representation dimensions are $2048$ and $512$ respectively, 
the variance score is in average very close 
to its upper bound. On the other hand, the trained AlexNet
(embedding dimension 9216) shows a strong decline from the randomly initialized version. 
This shows that the lack of 
variance in the embedding explanations is not merely due to the architecture inductive bias (as in the case for the random weights network),
but also on the representation learning itself. 
It is expected that such a high embedding size induces redundancy and therefore more entangled embeddings.
Accordingly, our experiment provides experimental evidence that the conciseness of a representation 
correlates with the desired property of disentanglement.

\subsection{Detecting Dataset Drift}
\label{ss:drift}

We can employ the noise score to detect the amount of drift 
from the training dataset. To test this, we train a simple autoencoder model on the 
\texttt{MNIST} dataset \cite{LeCun2005TheMD}.\footnote{MNIST dataset is available
under the terms of the Creative Commons Attribution-Share Alike 3.0 license.} The encoder network consists of two convolutional layers followed by
a linear layer. We refer to the SI for further details concerning hyperparameters,
training details and choice of architecture. We generate a drift by considering shifted data samples 
(normalized in range $[0,1]$) $\bx_{\lambda} = \bx + \lambda \by$, where $\bx$ is a sample from the original dataset, and the drift is generated by a normal 
distribution $\by \sim \cN(\mu=0,\sigma=1)$ and controlled by the parameter $\lambda$. 
We present in Figure \ref{f:dataset_drift} the relative difference of the noise scores for the 
second convolutional layer for various values of $\lambda$. Explicitly, this is defined as
$(N(\bx) - N(\bx_\lambda) )/ N(\bx)$. 
Here we denote $N(\bx) = N_{A_\Psi^p}=(\bx, \bx_{noise})$ for simplifying notation. 
Quantitatively, we obtain a Spearman coefficient of $\rho=0.988$ between the relative noise
scores and the corresponding drift parameter $\lambda$.
Our results confirm that the noise score
is negatively correlated with the shift amount. Hence, the noise score can be
used as a metric for dataset drift detection or, more generally, 
to quantify the representation quality degradation for datasets unseen during training time. 

\subsection{Enhancing Representations with Score Constraints}
\label{ss:enhance}

\begin{figure*}[t!]
\begin{minipage}{\textwidth}
  \begin{minipage}[t]{0.36\textwidth}
  \includegraphics[width=1.0\textwidth]{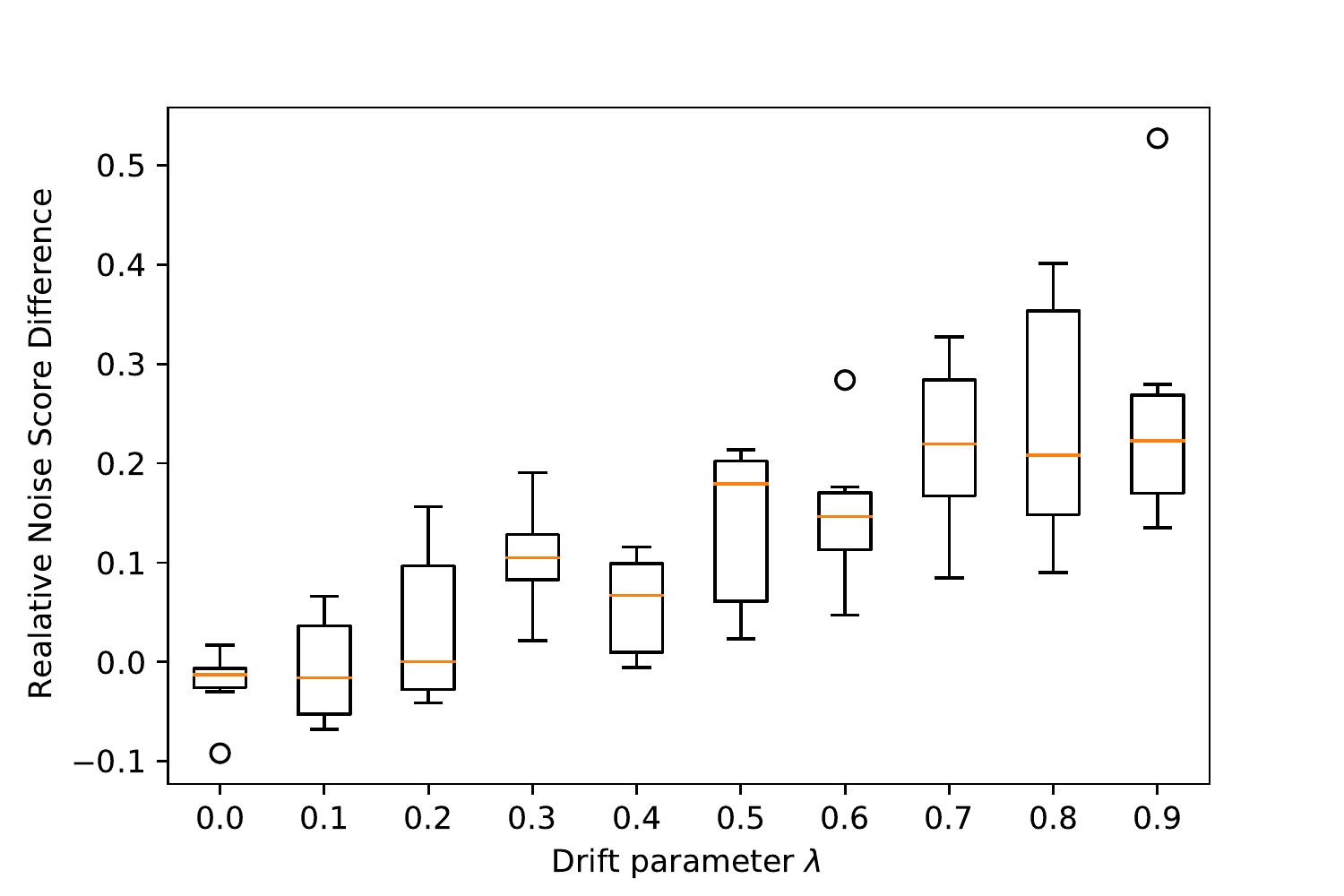}
  \caption{Correlation between noise score and dataset drift relized by introduction of 
  noise shift.}
  \label{f:dataset_drift}
  \end{minipage}
\hfill\vline\hfill
  \begin{minipage}[b]{0.61\textwidth}
    \centering
    \captionsetup{type=figure}
\captionof{table}{Classification accuracies for score-constrained embeddings on the \texttt{MNIST} dataset. The performances are averaged over 4 runs. NS = noise score,
VS = variance score.}
\label{t:constrs}
\begin{center}
\begin{small}
\resizebox{0.97\columnwidth}{!}{%
\begin{tabular}{lcc|c|c|c|c}
\toprule
Model & $\lambda_1$ & $\lambda_2$ &
dim$=2$ &
dim$=3$ & dim$=5$ & dim$=10$\\
\midrule
orig.   &0 & 0 & $88.3 \pm 0.7$ & $93.74\pm 0.3$  & $97.2\pm0.2$
& $97.7 \pm 0.2$\\
\midrule
NS &0.01 & 0  & $89.1\pm 0.5$ &   $94.3\pm 0.4$  & $97.4\pm 0.1$ 
& $97.8\pm 0.2$\\
NS &0.1 & 0  & $87.8 \pm 0.0$ &   $94.3\pm 0.2$ & $97.1\pm 0.3$ 
& $97.8\pm 0.3$\\
NS &1.0 & 0  & \boldmath{$89.3 \pm 0.5$} &   \boldmath{$95.1\pm 0.5$}
&$97.3\pm 0.1$ & $97.9\pm 0.1$\\
\midrule
VS  &0 & 0.01  & $88.6\pm 0.9$ &   $94.8\pm 0.8$ & \boldmath{$97.6\pm 0.1$} 
& \boldmath{$98.0\pm 0.2$}\\
VS  &0 & 0.1  & $88.5 \pm 1.0$ &   $94.5\pm 0.7$
& $97.4\pm 0.2$ &\boldmath{$98.0\pm 0.2$}\\
VS  &0 & 1.0  & $88.7 \pm 0.9$ &   $94.7\pm 0.6$ & $97.5\pm 0.2$
& $97.8\pm 0.2$\\
\midrule
NS+VS &0.1 & 0.1
& $88.5\pm 2.1$ &   $94.6\pm 0.8$  & $97.4 \pm 0.2$ & $97.8\pm 0.2$\\
\end{tabular}%
}
\end{small}
\end{center}
  \end{minipage}
\end{minipage}
\end{figure*}

We further test the usefulness of our scores by incorporating them as constraints during model 
training.
We train an autoencoder model $\Omega$ (with encoder $\Psi$)
on the \texttt{MNIST} dataset \cite{LeCun2005TheMD} with a linear
layer accessing the bottleneck embedding as a 10-class classifier. 
The training loss is therefore $\cL_{\text{train}}(\bx, y) = \cL_{\text{rec}}(\bx) 
+ \cL_{\text{class}}(\bx,y) - \lambda_1 N_{A_\Psi}(\bx, \by_{\text{noise}}) 
- \lambda_2 V_{A_\Psi}(\bx)$, where 
$\cL_{\text{rec}}(\bx) = |\bx - \Omega(\bx)|^2$ is the autoencoder reconstruction error and
$\cL_{\text{class}}$ is the classification loss (Cross Entropy Loss). The two coefficients
$\lambda_{1,2}$ control the ratios between the model loss and the score constraints. 
The encoder network $\Psi$ 
consists of two convolutional layers followed by a fully connected layer,
whose output is our target bottleneck representation. The scores are computed with respect
to the second convolutional layer.
In order not to affect excessively the training speed, we applied the scores
every 20 mini-batches. 
We did not perform any hyperparameter optimization: we trained the model for a fixed number of epochs (dependent on the bottleneck dimension), and we evaluate our results on a
held-out test set. 
We refer to the SI for further details concerning hyperparameters,
architecture and 
training details. 
We compare the
classification test performance of the model for various values of the coefficients $\lambda_{1,2}$, 
and we report our findings in Table \ref{t:constrs}.
We note that, overall, the score constraints led to representations that achieved a higher
test accuracy on the classification task.
We further notice that representation learning benefited from the variance score constraint
for higher embedding size: here, redundancy and entanglement affect the learning, and
forcing more concise representations reduces these undesirable effects.
On the contrary, the noise score constraint has bigger impact on the models with particularly
small bottlenecks, for which having a very expressive representation is key.
Finally, training including both scores outperforms the original model
in all the cases.

\section{Conclusions}
\label{s:concs}

In this work, we have proposed aggregation schemes to generalize gradient attribution methods to any two intermediate layers of a neural network. 
In particular, this framework can be used to probe the explanations of embeddings with respect to intermediate feature maps without the need for a downstream task or a direct mapping back to the input space. 
We derived two useful metrics:
the noise score can detect the informativeness of a representation, and this in turn correlates with downstream task performance; the variance score can be used to probe the disentanglement of an embedding. Both scores can be used as constraints for representation learning, and were shown in a proof-of-concept experiment to boost the performance of the learned representation on the downstream task of interest. 

This work is the first thorough study on how XAI techniques can be 
fruitfully implemented to evaluate and enhance representation learning
independently of a downstream task. 
It would be interesting
to apply our framework beyond the 
convolutional realm, for instance, to graph representation learning or neural machine translation. 
Our proposed variance aggregator can be seen as a first step towards explainability 
from the point of view of projection pursuit \cite{1672644}: it would be of interest to explore, 
both theoretically and experimentally, more general aggregation strategies based on higher momenta (e.g., lopsidedness).
Moreover, our proposed score constraints should be tested in more complex learning problems.
One limitation of our current approach is the increase of computational time due to the 
computation of the scores and the extra back-propagations during training. 
It would be important 
to derive ``lighter'' scores, which estimate \eqref{eq:noisescore} and \eqref{eq:varscore}
but with a lower memory and computation footprint. 
We plan to return to these issues in future work.

\bibliography{references.bib}

\clearpage

\appendix
\section{Variance Aggregator is sparser than Mean Absolute Aggregator}

We present a simple argument supporting our claim that the variance aggregator $\cA_{\text{var}}$ is 
sparser than the mean absolute aggregator $\cA_{\text{abs}}$.
Let $\cX$ be a multiset whose elements assume values in the range $[0,1]$.
In our case, this can be achieved by normalizing the multiset. 
Given the positiveness of $\cX$, the mean absolute aggregator simply reduces to the mean of 
$\cX$, i.e., $\cA_{\text{abs}} = \E[\cX]$.
Now we have that
\begin{align}
    \E[\cX]^2 = \E[\cX^2] - \Var[\cX] \leq \E[\cX] - \Var[\cX] \quad
    \Longrightarrow \quad \Var[\cX] \leq \E[\cX] (1 - \E[\cX])~,
\end{align}
where we used the fact that $\E[\cX^2]\leq\E[\cX]$ since $x^2\leq x$ for $x\in [0,1]$. Next, 
let us assume that $\Var[Z] > \E[Z]$, then
\begin{align}
    \E[\cX] < \E[\cX] (1 - \E[\cX]) \quad \Longrightarrow \quad \E[\cX] < 0~,
\end{align}
which violates the assumption of positiveness of $\cX$. Thus, we have shown that
$\cA_{\text{var}} = \Var[\cX] \leq \E[\cX] = \cA_{\text{abs}}$. 
In particular, the equality is achieved only for 
the multiset $\cX$ for which all elements equal zero, where $\cA_{\textit{var}} = \cA_{\text{abs}}=0$.
This simple argument shows that under some basic assumptions, given
the \textit{same} underlying distribution $\E[\Var Z] < \E[\E[Z]]$, and therefore 
the variance aggregation induces a sparser attribution map.

\section{Architecture, Hyperparameter and Training Strategy}

\subsection{Detecting Dataset Drift}

In Section 4.4 of the main text we report the results
of training an autoencoder model on the \texttt{MNIST} dataset,
showing that the noise score captures the effect of dataset drift.
The encoder model consists of $3$ convolutional layers of size $[8/5/1/2, 4/5/1/2, 2/5/1/2]$.\footnote{
The notation U/K/S/P completely defines a 
convolutional layer: U=number of units, K=kernel size, S=stride, P=padding.}
After each convolutional layer
we apply batch normalization and max pooling (kernel=2).
This produces a 32-dimensional bottleneck representation. 
The decoder network consists of 3 (transposed) convolutional layers of size $[4/5/1/2, 8/5/1/2, 1/5/1/2]$.
A sampling interpolation (scale factor=2)
precedes each convolutional layer, and we apply 
batch normalization after the first two convolutional layers.
We used ReLU as activation function, except for the final 
output of the decoder, where instead we apply a sigmoid function
to make sure the final output is in the range $[0,1]$.
We use Mean Squared Error (MSE) as our reconstruction loss. 
We use the Adam optimizer with a fixed learning rate
of $10^{-3}$, and we train for 5 epochs.

\subsection{Training with Scores as Constraints}

In section 4.5 of the main text we report our experiments 
regarding training an autoencoder model on the \texttt{MNIST} dataset with
our scores as constraints. 
The encoder model consists of $2$ convolutional layers of size
$[8/5/1/2, 4/5/1/2]$ 
followed by a fully connected layer of size $s$.
We apply batch normalization after each convolutional layer.
We conducted several 
experiments for $s=2,3,5,10$. We used softplus as the encoder activation function.
This is because when the 
score constraints enter the training loss, we need to
compute second derivatives with respect to the weights, which vanish if the activation functions are piece-wise linear. 
The classification branch of the network is a 10-dimensional fully connected layer accessing the bottleneck.
The decoder network consists of two fully
connected layers of size $[32, 28\times28]$. 
We used ReLU as a decoder activation function, and MSE as our reconstruction loss. 
We use the Adam optimizer with a fixed learning rate
of $10^{-3}$, and we train for a fixed number of epochs $n_{\text{ep}}(s)$
depending on the bottleneck size:
$n_{\text{ep}}(2) = 20$, 
$n_{\text{ep}}(3) = 15$, 
$n_{\text{ep}}(5) = 10$, and 
$n_{\text{ep}}(10) = 5$.

\subsection{Computation Time}

We report in Table \ref{t:comp-perf} the training time
for the autoencoder models, whose results we presented in 
Section 4.5 of the main text. When our
scores are added to the training as constraints, we observe an increasing computational time cost when the dimension of the target bottleneck embedding $\Psi$ increases. This is mainly due to the computation of
the scores themselves, as this involves a full backpropagation for 
each of the latent dimensions of $\Psi$. We observe that
the variance score has a lesser computational footprint than the 
noise score. One reason is that the former does not require the computation of attributions
for the noise-generated input. We expect that this alone introduces a factor of 2 on the computing time, and thus it cannot explain 
the full difference. We hypothesize that some inefficiency occurs
when we compute attributions for the noise input
\textit{after} having computed attributions for the meaningful input. 
In fact, in order to keep track of gradients, we set the option
\texttt{create\_graph=True} in the function 
\texttt{torch.autograd.grad} in PyTorch, and this might lead to an overall heavy computational derivative graph. 

\begin{table}[t!]
\caption{Execution time $[\frac{\text{seconds}}{\text{epoch}}]$ for the models reported in Section 4.5 of the main text. All the models were trained on a single NVIDIA Tesla V100-16GB GPU.}
\label{t:comp-perf}
\vskip 0.15in
\begin{center}
\begin{tabular}{l|cccc}
\toprule
Model &
dim$=2$ & dim$=3$ & dim$=5$ & dim$=10$\\
\midrule
orig. & 12 & 12 & 12 & 12\\
VS & 18 & 18 & 19 & 22\\
NS & 134 & 225 & 483 & 1569 \\
NS+VS & 135 & 222 & 500 & 1567
\end{tabular}%
\end{center}
\vskip -0.1in
\end{table}

\section{Further Embedding Visualization Examples}

\begin{figure}[t]
    \centering
    \includegraphics[width=0.85\textwidth]{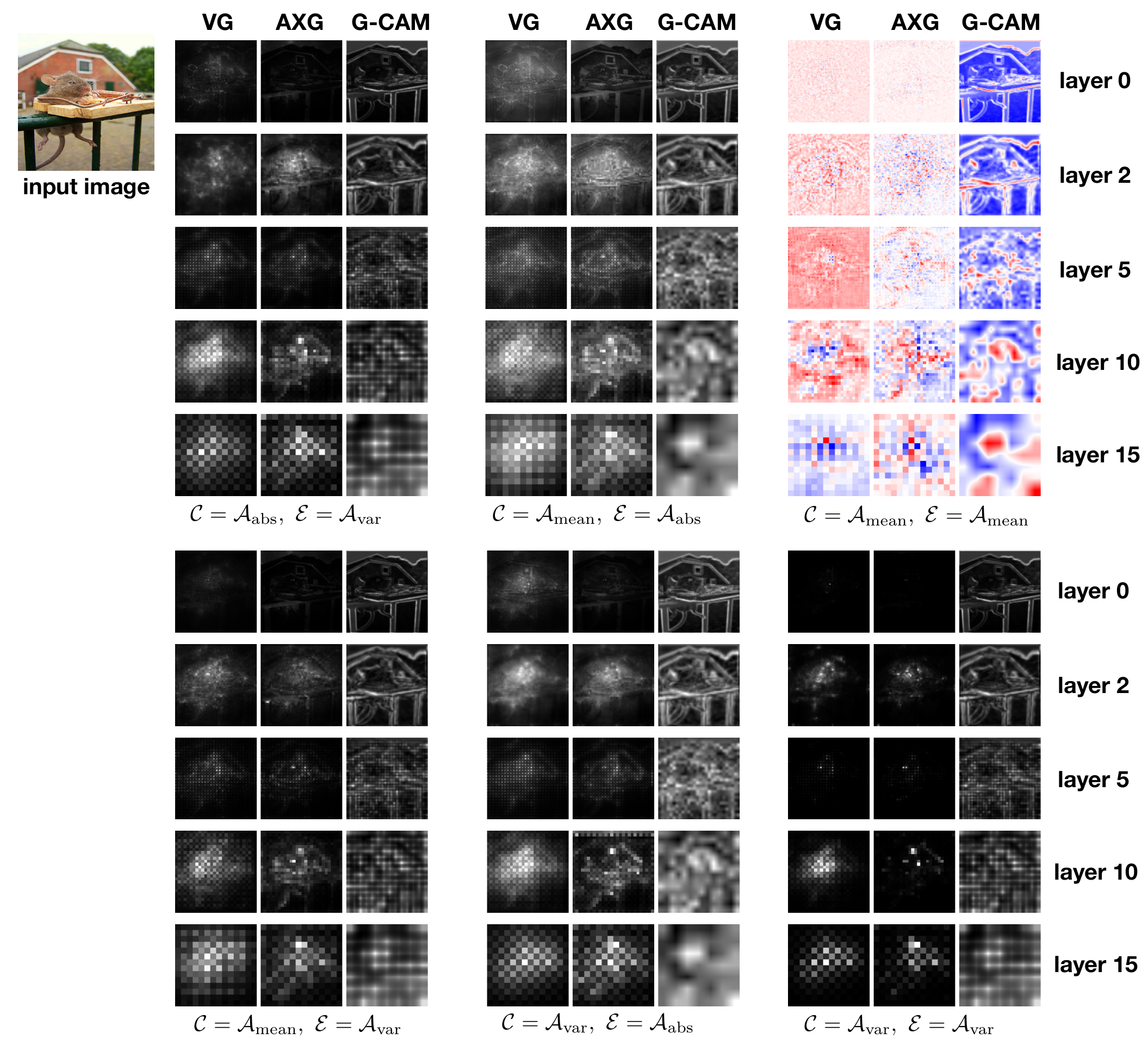}
    \includegraphics[width=0.85\textwidth]{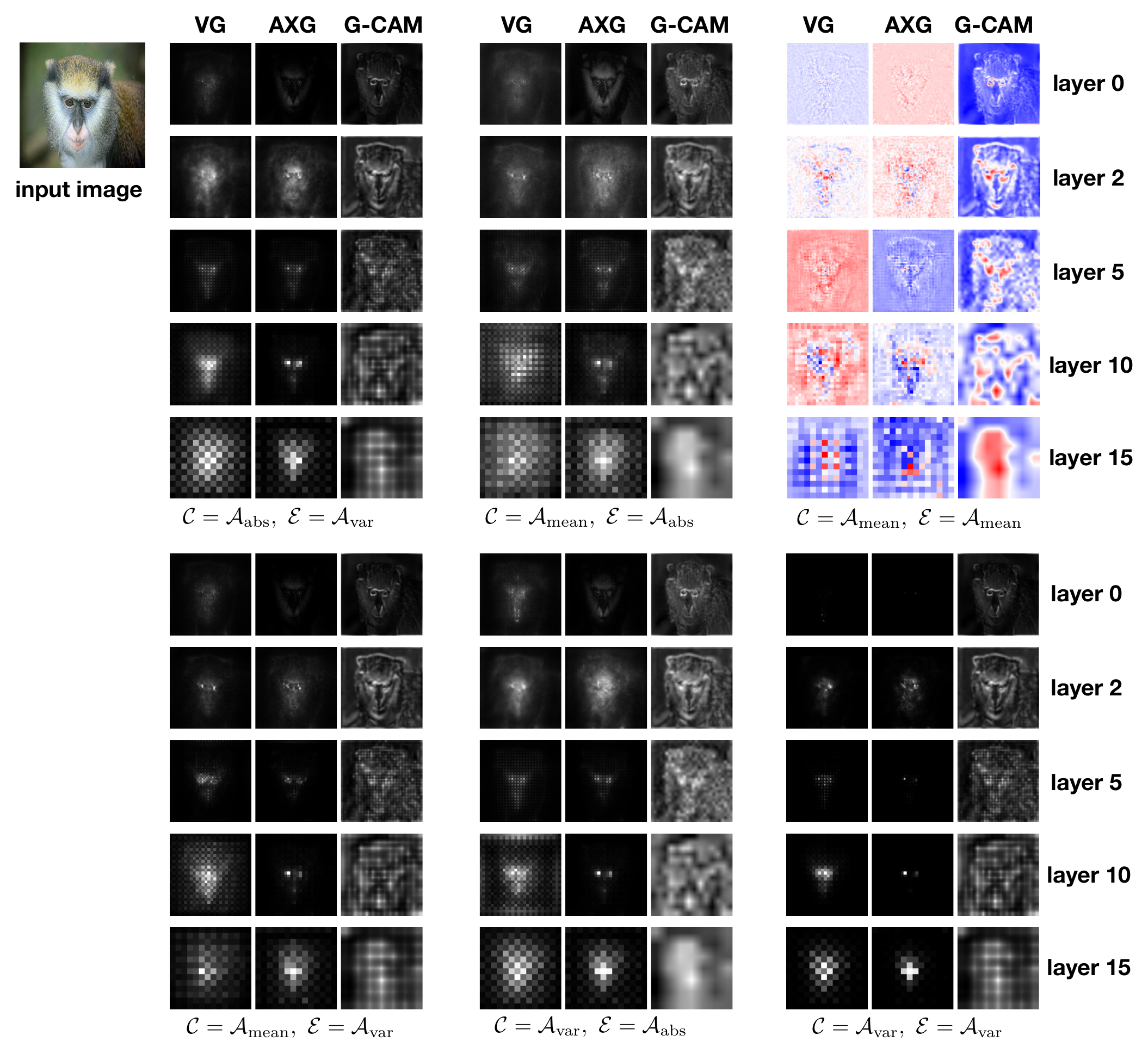}
    \caption{Embedding visualization for two examples for different attribution schemes and aggregation strategies.}
\label{f:attribs2}
\end{figure}

In the main text we restricted, for space reasons, to just a few examples regarding the possible combinations of 
attribution and aggregation strategies. In Figure \ref{f:attribs2} below we provide a more complete set of embedding visualization for the two input pictures
of Figure 2 of the main text. Specifically, for each example, we list,
for each of the attribution schemes (Vanilla Gradients, Activations$\times$Gradients, Grad-CAM), all possible combinations 
(for our channel and embedding aggregation maps $\cC$ and $\cE$,
respectively) of basic aggregation strategies 
we discussed in the main text ($\cA_{\text{mean}}$, $\cA_{\text{abs}}$
, $\cA_{\text{var}}$). Note that when $\cC = \cA_\text{var}, \cA_{\text{abs}}$, then $\cE = \cA_{\text{mean}}$ acts
equivalently to $\cE = \cA_{\text{abs}}$. 

\end{document}